\documentclass[final]{cvpr}

\usepackage{times}
\usepackage{epsfig}
\usepackage{graphicx}
\usepackage{amsmath}
\usepackage{amssymb}

\usepackage{amsfonts} 
\usepackage{amsmath}
\usepackage{capt-of,etoolbox}
\usepackage{multirow}
\usepackage{booktabs,tabularx}
\usepackage{diagbox}
\usepackage{enumitem}

\usepackage[pagebackref=true,breaklinks=true,colorlinks,bookmarks=false]{hyperref}

\makeatletter
\apptocmd\@maketitle{{\myfigure{}\par}}{}{}
\makeatother

\begin{document}
\newcommand\myfigure{
\centering
% \includegraphics[width=0.95\linewidth]{figures/intro.pdf}
%\captionof{figure}{(Best viewed in color) Our Facial Attribute Transformers (FAT) can transfer color and spatial attributes from the reference to the source. Spatial FAT is the extension of FAT that enables spatial transformation. The proposed facial attribute transformers can (1) more precisely estimate and apply colors to the source; (2) perform spatial transformation as shown in the column ``Spatial FAT''; (3) be generalized to other facial attribute transfer applications without modifications (see Sec.~\ref{sec:age-transfer} for details). FAT is also robust to different colors and poses, as shown in the second row.}
% \captionof{figure}{(Best viewed in color) Our \textbf{Facial Attribute Transformers (FAT)} can faithfully transfer detailed attributes (such as color and texture) from the reference to the source. Spatial FAT, an extension of FAT, further enables spatial transformation. The proposed facial attribute transformers exhibit clear advantages over previous methods in the following aspects: (1) \textbf{Fidelity}: FAT can transfer colors and details more precisely to the source (see the close-up of eyes and lips); (2) \textbf{Adaptability}: FAT can better handle the deformation and transfer of facial parts via spatial transformation (see the shape of the eyebrows in the last column); (3) \textbf{Generality}: FAT can be readily generalized to other facial attribute transfer tasks without modifications (see Sec.~\ref{sec:age-transfer} for details).}
% \vspace{0.2in}
% \label{fig:intro}
}

%%%%%%%%% TITLE
\title{Facial Attribute Transformers for Precise and Robust Makeup Transfer}

\author{Zhaoyi Wan\textsuperscript{\rm 1},
Haoran Chen\textsuperscript{\rm 2},
Jielei Zhang\textsuperscript{\rm 2},
Wentao Jiang\textsuperscript{\rm 3},
Cong Yao\textsuperscript{\rm 2},
Jiebo Luo\textsuperscript{\rm 1},\\
\textsuperscript{\rm 1}University of Rochester,
\textsuperscript{\rm 3}Beihang University,
\textsuperscript{\rm 2}Megvii\\
i@wanzy.me, jluo@cs.rochester.edu}

\maketitle
\vspace{-0.2in}

%%%%%%%%% ABSTRACT
\begin{abstract}
\vspace{-0.1in}
In this paper, we address the problem of makeup transfer, which aims at transplanting the makeup from the reference face to the source face while preserving the identity of the source. Existing makeup transfer methods have made notable progress in generating realistic makeup faces, but do not perform well in terms of color fidelity and spatial transformation. To tackle these issues, we propose a novel Facial Attribute Transformer (FAT) and its variant Spatial FAT for high-quality makeup transfer. Drawing inspirations from the Transformer in NLP, FAT is able to model the semantic correspondences and interactions between the source face and reference face, and then precisely estimate and transfer the facial attributes. To further facilitate shape deformation and transformation of facial parts, we also integrate thin plate splines (TPS) into FAT, thus creating  Spatial FAT, which is the first method that can transfer geometric attributes in addition to color and texture. Extensive qualitative and quantitative experiments demonstrate the effectiveness and superiority of our proposed FATs in the following aspects: (1) ensuring high-fidelity color transfer; (2) allowing for geometric transformation of facial parts; (3) handling facial variations (such as poses and shadows) and (4) supporting high-resolution face generation.
%Moreover, the proposed method can be easily generalized to other facial attribute transfer tasks, e.g., facial age editing (transfer), without modifications.
\end{abstract}

%%%%%%%%% BODY TEXT
\vspace{-0.2in}
\section{Introduction}

\begin{figure*}
    \centering
    \vspace{-0.3in}
    \includegraphics[width=0.95\linewidth]{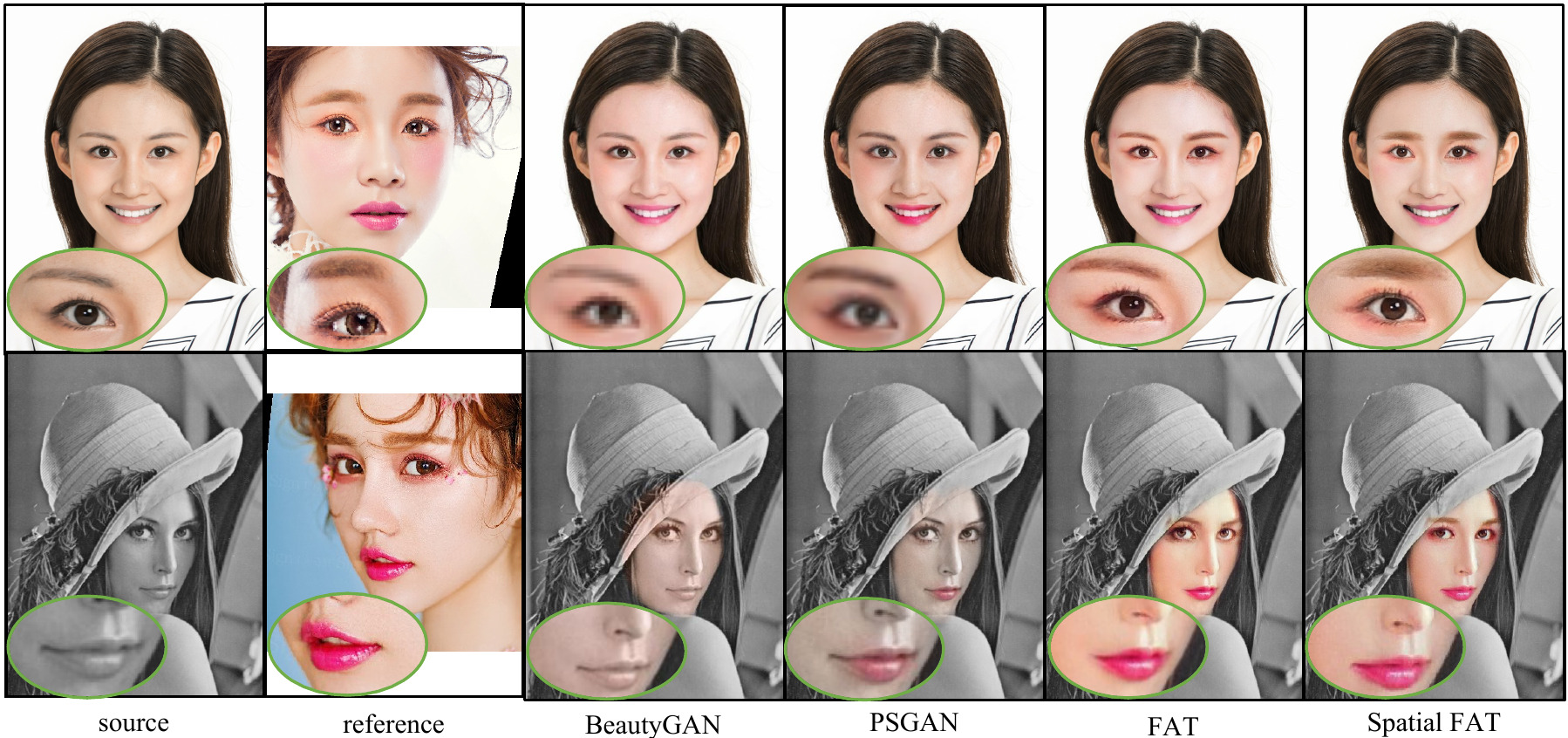}
    \captionof{figure}{(Best viewed in color) Our \textbf{Facial Attribute Transformer (FAT)} can faithfully transfer detailed attributes (such as color and texture) from the reference to the source. Spatial FAT, an extension of FAT, further enables spatial transformation. The proposed facial attribute transformers exhibit clear advantages over previous methods in the following aspects: (1) \textbf{Fidelity}: FAT can transfer colors and details more precisely to the source (see the close-up of eyes and lips); (2) \textbf{Adaptability}: FAT can better handle the deformation and transfer of facial parts via spatial transformation (see the shape of the eyebrows in the last column); (3) \textbf{Generality}: FAT can be readily generalized to other facial attribute transfer tasks without modifications (see Sec.~\ref{sec:age-transfer} for details).}
    \vspace{-0.2in}
    \label{fig:intro}
\end{figure*}

Makeup transfer has recently attracted much attention from the research community~\cite{beautygan, beautyglow, ladn}, since it possesses a series of technical challenges and has tremendous applicable value in numerous scenarios, for instance, online entertainment and cosmetics marketing.
As shown in Fig.~\ref{fig:intro}, the transfer procedure of makeup typically involves two face images, where the source provides facial identity and the reference exhibits makeup attributes, e.g., color, texture, and light and shade effect. The goal of makeup transfer is two-fold: Precisely morphing the given reference attributes into the source, while preserving the identity of the source.

Previous methods for makeup transfer generally work well and generate visually appealing makeup faces in various cases, such as large color ranges, arbitrary face poses and partial occlusions.
However, they might fall short when required to meet higher standards: \textit{high-fidelity color transplant and precise spatial transformation}.
As depicted in Fig.~\ref{fig:intro}, the results of existing algorithms (BeautyGAN~\cite{beautygan} and PSGAN~\cite{psgan}) are unsatisfactory when scrutinized with high standards.
Concretely, the colors of the lips are obviously different from that of the reference and the shapes of the eyebrows are kept unchanged.

%In this paper, we solve this problem in a new perspective, and propose novel Facial Attribute Transformers (FAT) to address these issues.
To enable high-quality makeup transfer, we propose a novel algorithm, named Facial Attribute Transformer (FAT), in this paper.
Taking advantage of the core idea of Transformer~\cite{transformer} in NLP, FAT adaptively models the semantic correspondence between the source face and the reference face.
It  adopts an attention mechanism and thus precisely generating the desired attributes.
In the framework of FAT, spatial transformations can be seamlessly integrated.
This extension of FAT, namely Spatial FAT, is the first makeup transfer method that integrates both color transplant and shape transformation.
It endows the system with the ability to transfer shape attributes, which is infeasible in previous algorithms.

The effectiveness of FATs (FAT and Spatial FAT) are verified through extensive qualitative and quantitative experiments (see Sec.~\ref{sec:experiments} for more details). We also design strategies for constructing ground truth for training FATs and producing high-resolution face images with details. Moreover, the proposed FATs can be easily generalized to other facial generation tasks that can be defined as color and spatial transformation of facial attributes (see Sec.~\ref{sec:age-transfer}).

The contributions of this paper can be summarized as follows:
\begin{itemize}[leftmargin=*]
\item We propose FATs (FAT and Spatial FAT), which promote the quality of makeup transfer to a new level: high-fidelity color transfer and precise shape transformation.
\item We specifically design a flexible ground truth generation strategy to provide proper guidance for the training of FATs and a powerful post-processing operation to produce high-resolution face images.
\item The proposed ideas are quite general and can be easily generalized to other facial attribute transfer applications, such as facial age transfer.
\end{itemize}

\vspace{-0.2in}
\section{Related Work}\label{sec:related-works}
\vspace{-0.1in}

% Jie Style Transfer
\noindent\textbf{Image Style Transfer and Makeup Transfer}
Image style transfer is a closely-related area to makeup transfer.
Starting from Gatys~\etal~\cite{synthsis}, image style transfer methods usually adopt iterative optimization~\cite{gatys2016preserving,risser2017stable} and feed-forward neural networks~\cite{frigo2016split,ulyanov1607instance} to render artistic effects on the produced images.
However, since style transfer methods do not take the strict semantic correspondence, \eg, the lips in the source face should correspond to the lips in the reference image, into consideration, the state-of-the-art style transfer algorithms~\cite{an2019ultrafast,lu2019optimal,wang2020collaborative,kim2020deformable,liu2020geometric} cannot be directly used to perform makeup transfer.
% Jie End

As for methods particularly towards makeup transfer, notable progress has been made to address these challenges.
LADN~\cite{ladn} proposes several hand-crafted local discriminators to drive the generator into reproducing color details.
BeautyGAN~\cite{beautygan} and its successors~\cite{psgan} adopt pseudo ground truth for each facial region using histogram equalization to solve pose inconsistency.
PairedCycleGAN~\cite{pairedcyclegan} proposes to guide makeup transfer using pseudo transferred images generated by blending the warped reference face to the source face.
PSGAN~\cite{psgan} proposes to disentangle the makeup into modulation matrices with an AMM module which enables shade-controllable and robust transfer.
Following these remarkable advances in makeup transfer, we further achieve a precise color and spatial transformation by proposing FAT in this paper.

\noindent\textbf{Transformers in GAN}
Transformer~\cite{transformer} was first devised in the natural language processing area to calculate the response at a position in a sequence as a weighted sum of the features at all positions by the self-attention mechanism.
Afterward, transformers and its variants are introduced into computer vision~\cite{dettransformer, imagewords} and GAN~\cite{sagan}.

The effort inspires new direction to image generation tasks~\cite{linestofacephoto, openpose}, and solves the problem in an attention-driven way.
Different from the self-attention paradigm of these applications, we develop the concept of self-attention into a mutual-correspondence manner and devise FAT for aligning two faces whose attributes are supposed to be precisely estimated and transferred.

\section{A Brief Review of Makeup Transfer}\label{sec:revisit}

\begin{figure}
    \centering
    \includegraphics[width=0.95\linewidth]{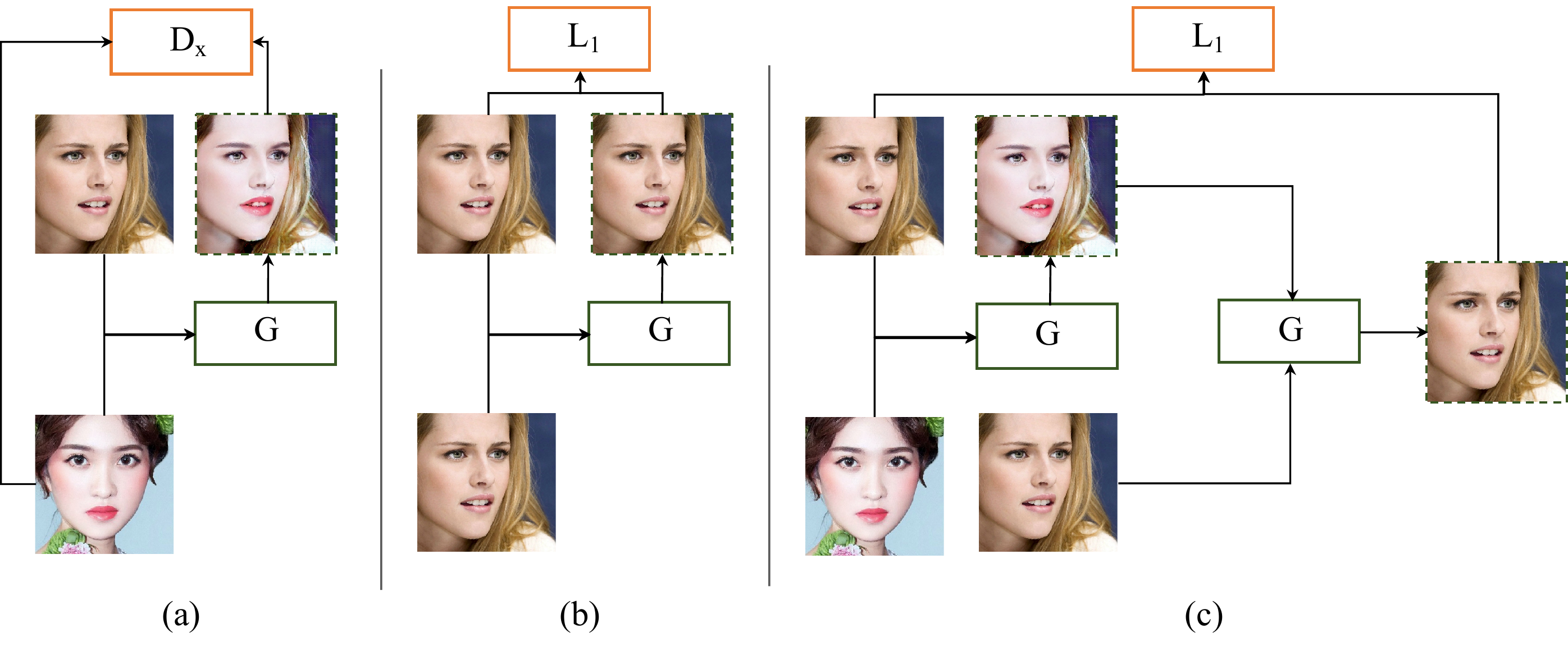}
    \caption{The GAN training of makeup transfer. $G$ is short for the generator, and $D_x$ is the discriminator for reference. $L_1$ is the Manhattan distance that performs as a criterion. The generator output is marked with dash borders. Note the reference and source are symmetrically exchanged in each pass during training.}
    \label{fig:gan}
    \vspace{-0.2in}
\end{figure}

We briefly review makeup transfer in this section, introducing crucial components for the implementation of a makeup transfer model.
Makeup transfer (and other facial attribute transfer) is typically modeled by training a mapping which transfers attributes from the reference to the source: $G: \{x, y\} \rightarrow z$ ~\cite{psgan}, where $x$ and $y$ are samples from the source and reference space, respectively.
The combination of the desired attributes of reference and identity of the source is referred to as $z$.

\subsection{Model Formulation}

Specifically, the basis of the learned mapping is two kinds of functions, \textit{transformation function} $\Gamma(\cdot)$ and \textit{estimation function} $E(\cdot)$:
\begin{equation}\label{eq:gamma}
        \hat{z} = \Gamma(\hat{x}, E(\hat{y})),
\end{equation}
where $\Gamma(\cdot)$ applies the attributes estimated by $E(\cdot)$ to $x$.
Note that the transformation can be and is usually performed on high-dimensional features.
Thus, we use $\hat{x}$ and $\hat{y}$ in Eq.~\ref{eq:gamma}.
Literature presents different preferences in choosing $\Gamma(\cdot)$ and $E(\cdot)$, such as MLP and concatenation in LADN~\cite{ladn}, convolution and linear in PSGAN~\cite{psgan}.

Existing transformation methods only consider the transformation in color space, although spatial attributes are also important in make transfer.
Our facial attribute transformer is designed to be compatible with spatial transformations, thus making spatial attributes transformable.

\subsection{Training with GANs}

Most of the existing methods for makeup transfer are based on Generative Adversarial Networks (GANs).
A consensus the community has reached is that only unpaired training data is used for training makeup transfer.
Instead of collecting facial images with and without makeup of the same person, the training strategy from Cycle-GAN~\cite{cyclegan} is used as the common practice for makeup transfer to train generators from face images of different people.

The training procedure of Cycle-GAN-based methods is illustrated in Fig.~\ref{fig:gan}.
The adversarial generation is additionally supervised by the consistency losses.
The figure omits the symmetrical forward where the role of source and reference images are exchanged.
Specifically, the loss for GAN training of the generator is an aggregation of adversarial loss, consistency loss, and perceptual loss.
\begin{equation}
    \begin{gathered}
    \begin{aligned} J_{D}^{adv} &= - \mathbb{E}_{x \sim \mathcal{P}_{X}}\left[\log D_{X}(x)\right] - \mathbb{E}_{y \sim \mathcal{P}_{Y}}\left[\log D_{Y}(y)\right] \\ &- \mathbb{E}_{x \sim \mathcal{P}_{X}, y \sim \mathcal{P}_{Y}}\left[\log \left(1-D_{X}(G(y, x))\right)\right] \\ &- \mathbb{E}_{x \sim \mathcal{P}_{X}, y \sim \mathcal{P}_{Y}}\left[\log \left(1-D_{Y}(G(x, y))\right)\right] 
        \end{aligned} \\
        \begin{aligned}
            J_G = \lambda_{adv} J_G^{adv} + \lambda_{cyc} J_G^{cyc} + \lambda_{per} J_G^{per},
        \end{aligned}
    \end{gathered} 
\end{equation}
where
\begin{equation}
    \begin{gathered}
        \begin{aligned} J_{G}^{adv} = &- \mathbb{E}_{x \sim \mathcal{P}_{X}, y \sim \mathcal{P}_{Y}}\left[\log \left(D_{X}(G(y, x))\right)\right] \\ &- \mathbb{E}_{x \sim \mathcal{P}_{X}, y \sim \mathcal{P}_{Y}}\left[\log \left(D_{Y}(G(x, y))\right)\right] \\
    \end{aligned}\\
    \begin{aligned}
     J_G^{cyc} &=\mathbb{E}_{x \sim \mathcal{P}_{X}, y \sim \mathcal{P}_{Y}}\left[ \left\| G(G(x, y), x) - x \right\|_{1} \right]  \\ &+\mathbb{E}_{x \sim \mathcal{P}_{X}, y \sim \mathcal{P}_{Y}}\left[ \left\| G(G(y, x), y) - y \right\|_{1} \right]
    \end{aligned}\\
    \begin{aligned}
     J_G^{per} &=\mathbb{E}_{x \sim \mathcal{P}_{X}, y \sim \mathcal{P}_{Y}}\left[ \left\| F_l(G(x, y)) - F_l(x) \right\|_{2} \right]  \\ &+\mathbb{E}_{x \sim \mathcal{P}_{X}, y \sim \mathcal{P}_{Y}}\left[ \left\| F_l(G(y, x)) - F_l(y) \right\|_{2} \right].
    \end{aligned}
    \end{gathered}
\end{equation}

In each pass of the training of the generator, the symmetrical loss functions are added together to compute the gradient.
We don't go deep into the loss formulation details of Cycle-GANs and refer readers not familiar with it to Appendix A.

\subsection{Makeup Loss with Pseudo Ground Truth}

\begin{figure}[tbp]
    \centering
    \includegraphics[width=0.95\linewidth]{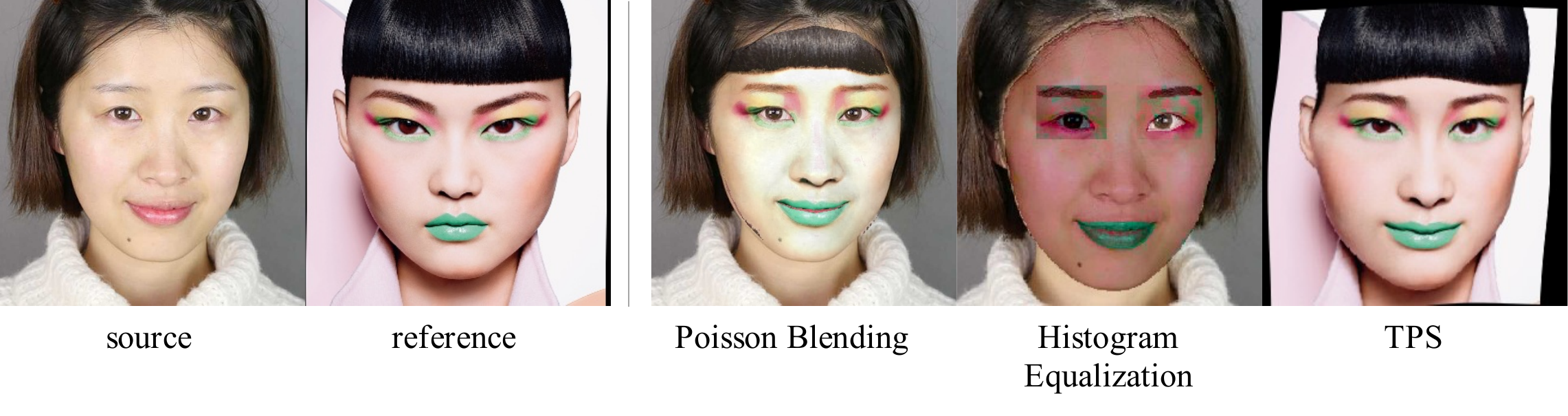}
    \caption{Different options for pseudo ground truth generation.}
    \label{fig:sup}
    \vspace{-0.2in}
\end{figure}

The above-mentioned training strategy drives the generator to produce realistic images with the makeup.
Nevertheless, the goal of makeup transfer is more than producing realistic facial images. To reconstruct the makeup attributes of the reference on the source face, extra supervision with pseudo ground truth (pseudo GT or $PGT$) is introduced:
\begin{equation}
    \begin{aligned}
        J_G^{make} &= \mathbb{E}_{x \sim \mathcal{P}_{X}, y \sim \mathcal{P}_{Y}}\left[ \left\| G(x, y) - PGT(x, y) \right\|_{2} \right]  \\ &+\mathbb{E}_{x \sim \mathcal{P}_{X}, y \sim \mathcal{P}_{Y}}\left[ \left\| G(y, x) - PGT(y, x) \right\|_{2} \right].
    \end{aligned}
\end{equation}

% Since we know that the data for makeup transfer is unpaired, the mainstream solution is to generate pseudo ground truth for training.
Literate has shown at least two representative families of methods for pseudo ground truth generation.
As shown in Fig.~\ref{fig:sup}, blending~\cite{pairedcyclegan, ladn} and histogram equalization~\cite{psgan, beautygan} are typically used for GT generation.
The blending strategy recovers the color and context of reference images that are at similar alignment with the source but is sensitive to misalignment between the source and reference.
Histogram equalization adjusts the color distribution of the source images, thus being robust to different poses and expressions in the reference.
However, it suffers from shadows and extreme colors in the reference due to the loss of spatial distribution.

Although the generated pseudo GT is different from generated images in quality, it provides coarse yet sufficient guidance in complement to the GAN training.
On the other hand, most of the existing methods follow the concept where the pseudo GT demonstrates a transformation from the source to the reference.
In this paradigm, the demonstrated transformation is usually imprecise in desired attributes and causes sub-optimal transformation.
Alternatively, we propose a novel GT generation which aligns the pose and orientation of the reference face to the source face, creating pseudo GT with precise color.

\section{Methodology}

\begin{figure}[tp]
    \vspace{-0.2in}
    \centering
    \includegraphics[width=0.75\linewidth]{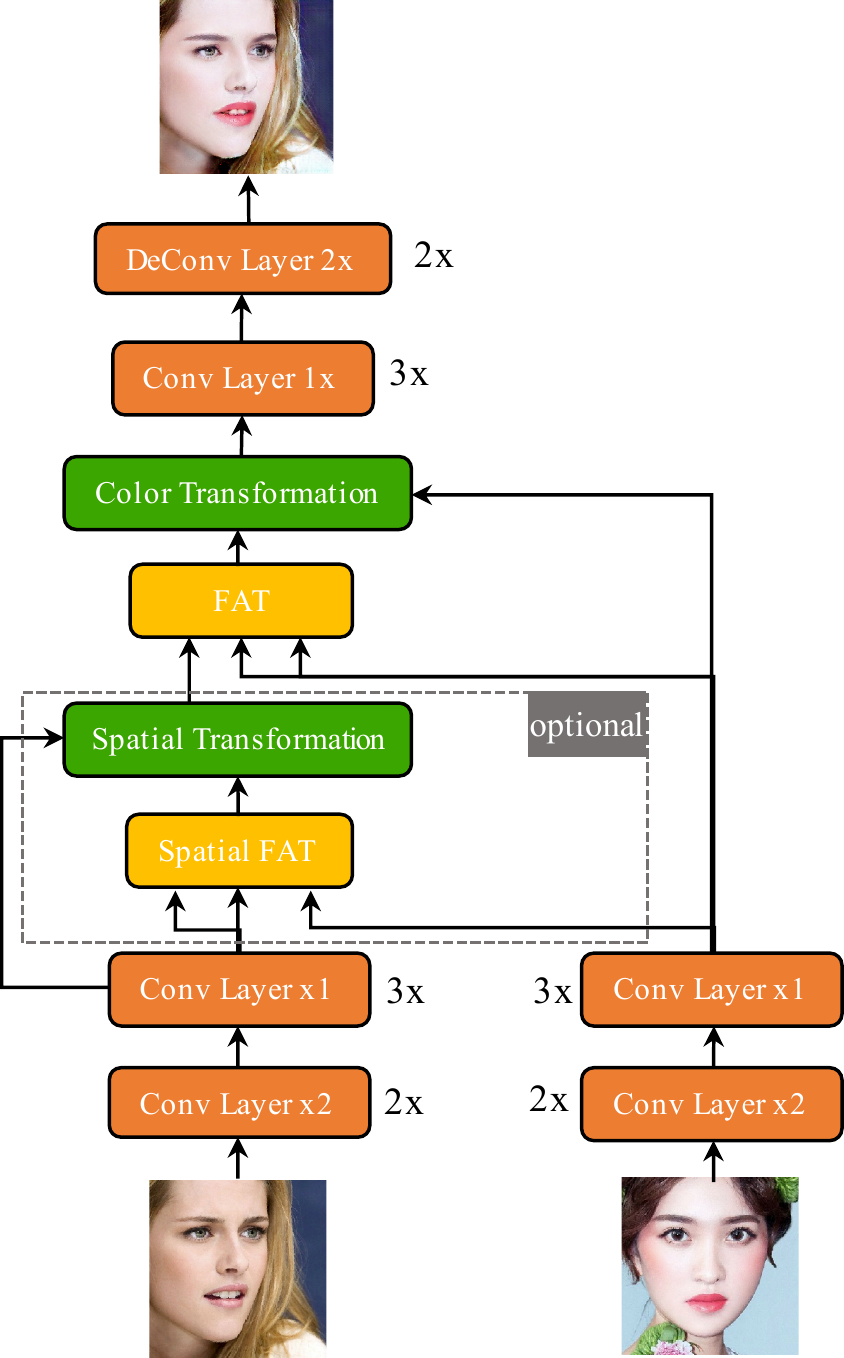}
    \caption{Overall structure of our proposed generator, where FAT and Spatial FAT are applied for color transformation and spatial transformation, respectively.
    Each ``Conv Layer'' is composed of a convolution, instance normalization, and ReLU activation.
    ``x1, x2'' are the stride of (de-) convolutional layers, and ``2x, 3x'' indicates repeatedly stacked layers.}
    \label{fig:net}
    \vspace{-0.2in}
\end{figure}

So far we have revealed the main limitations of existing makeup transfer methods: the absence of the ability to perform spatial transformations, and imprecise color generation. In this section, we present our novel approaches to address these issues. Following \cite{cyclegan, psgan, pix2pix}, we use an encoder-decoder network as the basic architecture of our models as shown in Fig.~\ref{fig:net}. 

\subsection{Facial Attribute Transformer}

As shown in Sec.~\ref{sec:related-works}, attention mechanisms are introduced to the face generation and show promising improvements in the robustness of generative models. However, the existing attention module relies on similarity measurements on each part of the faces. It is limited by the absence of spatial modeling and suffers from low efficiency. For example, PSGAN~\cite{psgan} is 10x slower than its baseline due to the cost of a sequential computation of attention at each face part.

Inspired by the self-attention mechanism in Transformer~\cite{transformer} in NLP, we devise a facial attribute transformer to model the correspondence between the source face and reference face. Given the features extracted from the source and reference images, $\hat{x}$ and $\hat{y}$, the attention matrix is computed as:
\begin{equation}\label{eq:atten}
    A(\hat{x}, \hat{y}) = Softmax(\frac{\hat{x}W_x(\hat{y}W_y)^T}{\sqrt{d}}),
\end{equation}
where $W_x$ and $W_y$ are learnable parameters and $d$ is the dimension of features. Naturally, $\hat{x}$ and $\hat{y}$ are flatten along the height and width dsimension to fit the dot-product attention.
With this formula, attention of $k$ parts, e.g., eyes and lips, can be efficiently performed in parallel:
\begin{equation}
    \hat{A}(\hat{x}, \hat{y}) = Concatnate(A_1(\hat{x}, \hat{y}), \dots, A_k(\hat{x}, \hat{y}))W_o.
\end{equation}
Alternative to the self-attention mechanism~\cite{transformer, sagan}, the attention matrix is applied to the attributes distilled from $\hat{y}$.
Let $\gamma_y$ be the attributes directly estimated from $\hat{y}$ using convolution layers. 
It consequently maintains spatial correspondence with $\hat{y}$.
Then applying the attention matrix $\hat{A}$ transfers $\gamma_y$ to be corresponded to the spatial distribution of $\hat{x}$.
\begin{equation}
\begin{split}
    \gamma_x & = \hat{A}\gamma_y,\\
    \hat{x}' & = \Gamma(\hat{x}, \gamma_x).
\end{split}
\end{equation}
Notably $\gamma_x$ and $\gamma_y$ are high-dimension representations of attributes that maintains richer information than a single-dimensional representation.

As introduced in Sec.~\ref{sec:revisit} , a specific transformation function $\Gamma$ is used to morph the distilled attributes to the source.
We use 
\begin{equation}
    \Gamma_c(\hat{x}, \gamma_x) = \gamma_x \begin{bmatrix}
    \hat{x} \\
    I
    \end{bmatrix}
\end{equation}
for color transformation, where $I$ is the identity matrix with the same shape with $\hat{x}$. Source features $\hat{x}$ are thus linearly transferred to reconstruct the color of reference images.

\begin{figure}
    \vspace{-0.2in}
    \centering
    \includegraphics[width=0.95\linewidth]{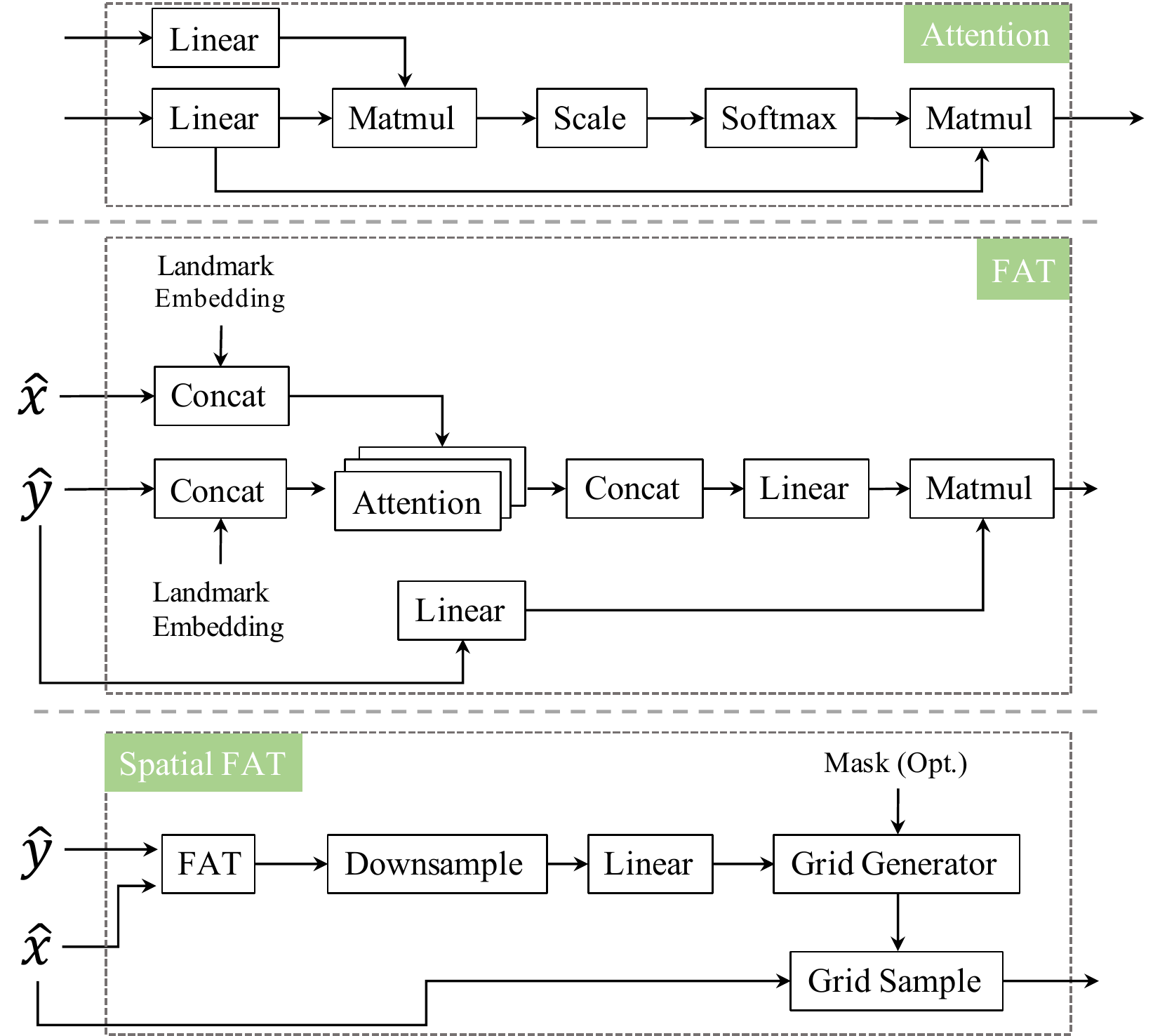}
    \caption{Structure of FAT and spatial FAT. ``Scale'' uses the factor $\sqrt{d}$ as shown in Eq.~\ref{eq:atten}.}
    \label{fig:fat}
    \vspace{-0.2in}
\end{figure}

\subsection{Spatial FAT}

FAT aligns two faces and is designed for linear transformation of attributes, such as color and texture.
To further enable spatial transformation, which is crucial for many facial attribute transfer applications, we extend FAT into Spatial FAT  in this section.
Spatial FAT is inspired by the Spatial Transformer Networks~\cite{stn} (STN) and similarly predicts a spatial transformation that is applied to the features.

Specifically, we introduce Thin Plate Spine (TPS), the transformation Spatial FAT predicted to enable spatial attributes transfer.
TPS is widely used in digital image processing such as face morphing~\cite{tpsmorphing} and shape matching~\cite{rare}.
It is more flexible than rigid transformations such as affine transformation and can achieve local spatial transformation but minimizing the global distortion~\cite{tps}.
Specifically, TPS is determined by an array of $K$ \textit{control points} whose coordinates in the original space and the desired target space are presented as $C = [c_1, c_2, \dots, c_K]$ and $C' = [c_1', c_2', \dots, c_K']$, respectively.
As shown in Fig.~\ref{fig:pseudogt}, TPS warps the original image to fit $C$ into $C'$.

Different from STN, we use lower-resolution features to predict the TPS transformation to grasp the spatial differences in Spatial FAT.
Its structure is illustrated in Fig.~\ref{fig:fat}.
Features $\hat{y}$ from the reference are initially aligned using a regular FAT, and $\hat{x}$ and $\hat{y}$ are passed to FAT in a reversed order for this purpose.
Then each pixel position in the down-sampled feature map is regarded as a control point, resulting in $K=h\times w$ points. 
A convolutional layer with a Tanh activation function is applied to predict the target control points, which are normalized to $[-1, 1]$.
Similar to STN, we use a sample grid computed from the control points to implement the warp of TPS transformation.
The details about grid construction are provided in Appendix B.

\subsection{Landmark Embedding and Face Parsing}
In the vanilla transformer~\cite{transformer}, position embedding is adopted to capture spatial information.
In our FAT, we consider the landmarks of faces laying pivotal spatial guidance for attribute discovering and face generation.
Therefore, we introduce Landmark Embedding (LE) as a replacement to the position embedding in vanilla transformers.
Given the facial images with $N$ landmark points, we compute the distance from each pixel to all the landmark points as an embedding:
\begin{equation}
    LE_{cor} = [cor - L_i]^T, i = 1, 2, \dots, N
\end{equation}
where $cor$ is a coordinate inside the image.
$LE$ is then flattened, normalized by its 2-norm, and used in FAT to enhance visual features, as shown in Fig.~\ref{fig:fat}.

Following PSGAN~\cite{psgan}, we use face parsing results to control makeup transfer on particular parts of faces.
For example, the grid generated by Spatial FAT can be masked by the face parsing masks that only desired parts of the face will be spatially warped.
Thus, partial makeup shown in \cite{psgan} is also theoretically supported in our work.

\subsection{Ground Truth Generation}\label{sec:label-generation}

\begin{figure}
    \includegraphics[width=1.0\linewidth]{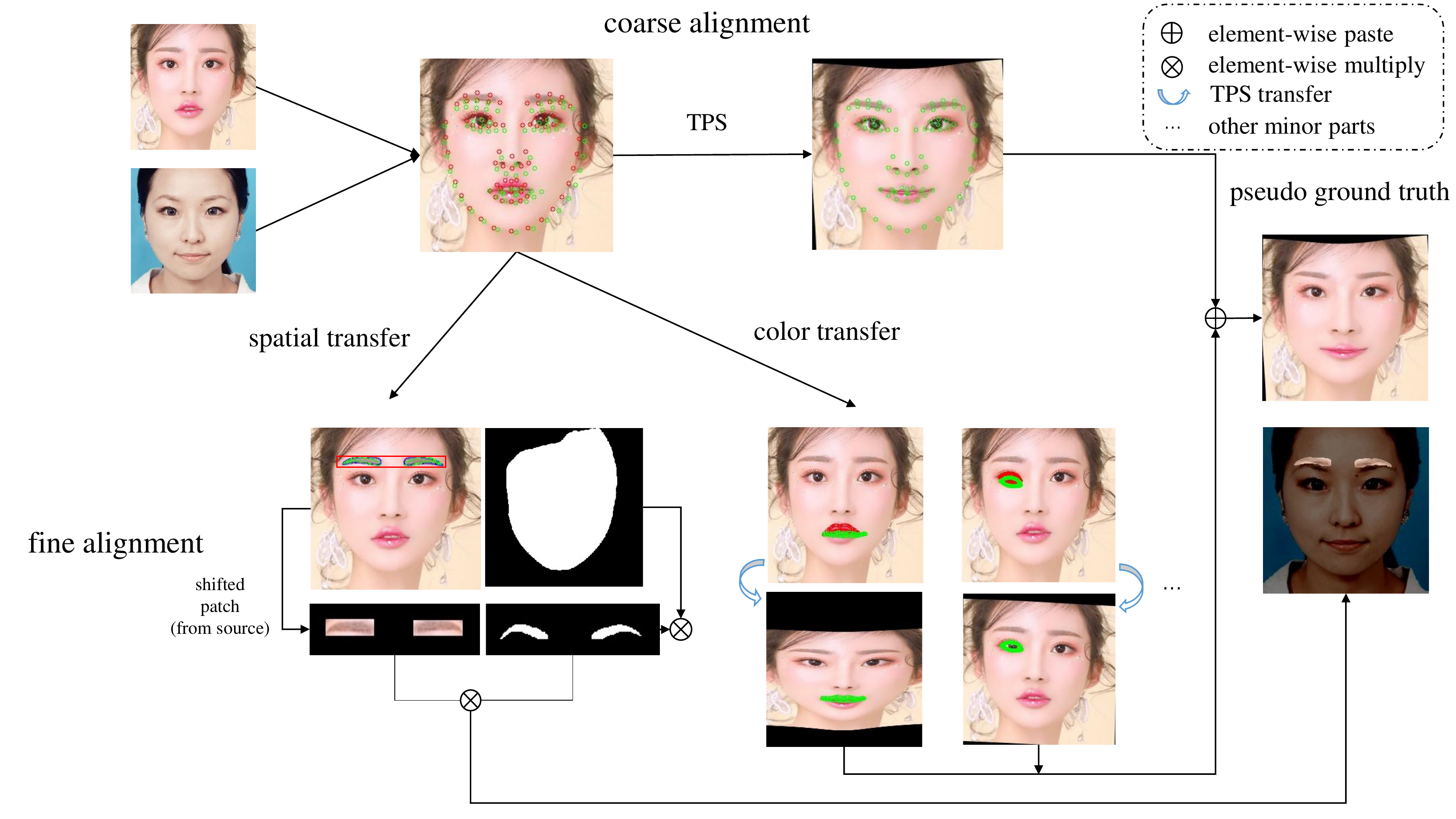}
    \caption{Illustration of the proposed pseudo ground truth generation method. The proposed pseudo ground truth generation follows a coarse to fine manner.
    The reference face is first warped to match the coarse shape of the source, then minor parts such as lips and eyebrows are tuned by particular TPS transformation.}
    \label{fig:pseudogt}
    \vspace{-0.2in}
\end{figure}

As stated in Sec.~\ref{sec:revisit}, the GAN-based training strategy supervises makeup transfer to be \textit{realistic}.
As a complement, pseudo ground truth (GT) that serves as coarse guidance, enforces the transformation to be \textit{accurate}.
We show our pseudo ground truth generation in Fig.~\ref{fig:pseudogt}.
Our strategy is based on TPS transformation and produces color and spatial transformation ground truth separately.

\noindent\textbf{GT for Color Transformation}
Different from most existing approaches, we generate color transformation GT by applying TPS transformation on the reference image to fit the face of the source image.
It first warps the face according to coarse landmarks, and tune each face parts using corresponding points.
Specifically, lips, eyebrows, and eyes are separately warped along with the face parsing mask, which is used to paste the corresponding regions back to the coarse GT.
As Fig.~\ref{fig:pseudogt} shows, the GT generated by TPS transformation is not absolutely aligned with the source image due to the limitation of the preciseness of landmarks.
However, we favor this approach that generates GT from the reference in consideration of two reasons.
First, the quality of generated images is well-supervised by GAN losses, but the accuracy of transformations is not under precise guidance.
Second, the GT generation from reference conserves detailed facial attributes and thus, can be generalized to wider application scenarios that conventional GT generation usually fails to deal with.
The results shown in Sec.~\ref{sec:experiments} also validate our design.

\noindent\textbf{GT for Spatial Transformation}
\begin{figure*}[htpb]
    \vspace{-0.2in}
    \centering
    \includegraphics[width=0.95\textwidth]{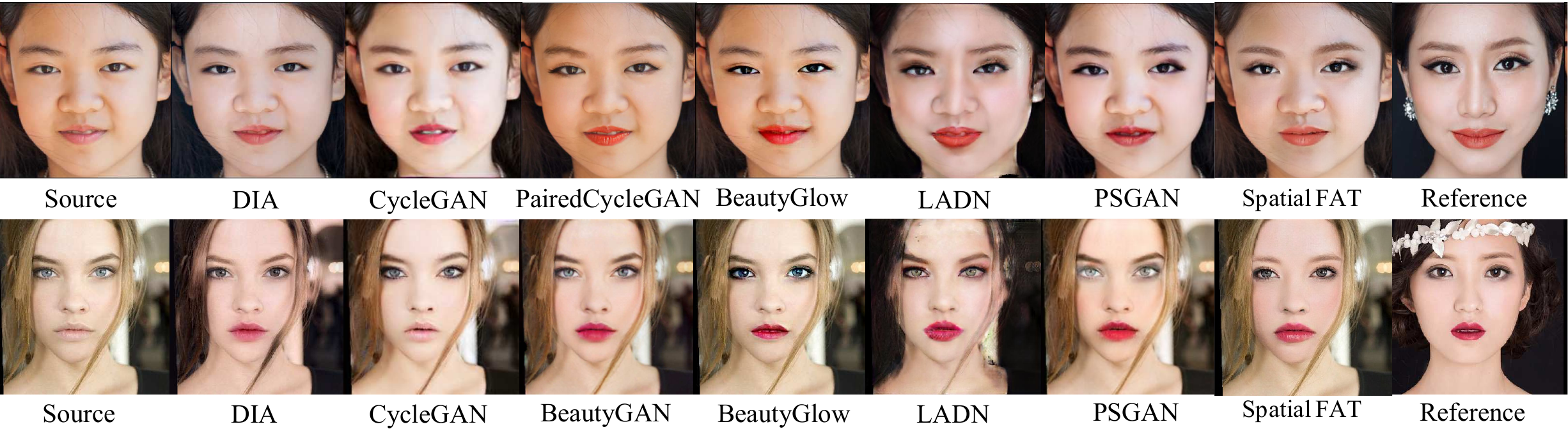}
    \caption{Comparison with state of the art methods. Spatial FAT precisely transfers the colors of reference, resulting in ideal reconstruction of makeup.
    Furthermore, it is the first method that can transfer the shape of eyebrows.}
    \label{fig:general-comparison}
    \vspace{-0.2in}
\end{figure*}
For a proper spatial transformation, we aim at transferring the shape of a part in reference face, but preserving the location in the source face.
For example, the straight eyebrows of reference should be transferred, to the position of the crescent eyebrows of source in Fig~\ref{fig:pseudogt}.
More formally, given two regions on the source and reference which are surrounded by $N_p$ points $P = \{P^1, \dots, P^{N_p}\}$ and $Q = \{Q^1, \dots, Q^{N_p}\}$, respectively, the fundamental goal of GT generation is to find a \textit{shift transformation} parameterred by $\Delta_Q$ that minimizes the overall distance between the source points and the shifted reference points $\sum_{i=1}^{N_p}Dis(P^i, Q^i - \Delta_Q)$, where $Dis(\cdot)$ is a distance measure.
Taking the Euclidean distance, we can resolve the min-distance shift by
\begin{equation}
\begin{split}
    &\frac{\partial \sum_{i=1}^{N_p}\|P^i - Q^i + \Delta_Q\|^2}{\partial \Delta_Q} = 0\\
    \Rightarrow & \Delta_Q = \frac{\sum_{i=1}^{N_p}{(Q^i - P^i)}}{N_p}
\end{split}
\end{equation}
The generation procedure is illustrated in Fig~\ref{fig:pseudogt} and our implementation of pseudo GT generation is provided in the supplementary materials.

Note the presented spatial GT generation is specially designed for spatial attributes transformation and supposed to be used with Spatial FAT (see ablation study in Sec.~\ref{sec:ablation}).
Enhanced by the proper supervision, Spatial FAT can generate realistic faces where spatial attributes are also transferred, as shown in Sec ~\ref{sec:experiments}.

\section{Experiments}  \label{sec:experiments}

\begin{table}[tp]
\centering
\begin{tabularx}{1.0\linewidth}{@{}l*{6}X@{}}
\toprule
\multirow{2}{*}{Method}     & \multicolumn{3}{c}{Functionality}   &  \multirow{2}{*}{Robust}  \\
                            & Color & Texture & Shape &  \\ \midrule
BGAN~\cite{beautygan}       &  $\checkmark$ & $\times$ & $\times$ & $\times$     \\
PCGAN~\cite{pairedcyclegan} &  $\checkmark$ & $\times$ & $\times$ & $\times$     \\
BGlow~\cite{beautyglow}     &  $\checkmark$ & $\times$ & $\times$ & $\times$      \\
LADN~\cite{ladn}            &  $\checkmark$ & $\checkmark$ & $\times$ & $\times$   \\
PSGAN~\cite{psgan}          &  $\checkmark$ & $\times$ & $\times$ & $\checkmark$   \\
FAT (ours)                  &  $\checkmark$ & $\checkmark$ & $\checkmark$ & $\checkmark$    \\
   \bottomrule
\end{tabularx}
\caption{Functionality analysis of the makeup transfer methods. By referring to a method ``robust'', we indicate it is robust to facial orientation and shadows.}
\label{tab:functionality}
\end{table}

We conduct extensive experiments to validate the effectiveness of our proposed modules.
Basically, we use the MT~\cite{beautygan} dataset for the training of makeup transfer.  MT contains 3,834 images where 2,719 images are with makeup.
The test data from \cite{psgan}, M-Wild, is also introduced for comparison.

\subsection{Implementation Details}

All models are with a fixed number of attentions $k=2$ and trained with a learning rate of $2e^{-4}$ and the Adam~\cite{adam} optimizer.
For makeup transfer, we train our models with the MT dataset for 50 epochs.
Although the experiments are conducted with $256\times256$ resolution of focused facial images, higher resolution images can be produced using our post processing strategy described in Sec.~\ref{sec:post-processing}.
Our methods are implemented with PyTorch~\cite{paszke2019pytorch} and will be publicly available with the acceptance of the paper.

\subsection{General Comparison}

We start our experiments with a general qualitative comparison with the state-of-the-art methods, including image style transfer methods DIA~\cite{dia}, CycleGAN~\cite{cyclegan}, PairedCycleGAN~\cite{pairedcyclegan}, and makeup transfer methods BeautyGlow~\cite{beautyglow}, LADN~\cite{ladn}, PSGAN~\cite{psgan}.
As the implementation of BeautyGlow and PairedCycleGAN is not publicly released, we follow BeautyGAN~\cite{beautygan} and crop the results from their paper.

As shown in Fig.~\ref{fig:general-comparison}, most of the existing methods can generate makeup faces and keep the identity of source images.
Among these methods, DIA achieves better color reconstruction in lips and eyebrows, but suffers from unnatural color in the face and missing of eye shadows.
More recent methods such as BeautyGlow and PSGAN can - although the color is not precise - transfer eye shadows.
In comparison, the first row of Fig.~\ref{fig:general-comparison} demonstrates the proposed FAT can precisely transfer makeup colors with realistic results, where the identity and light on the source face are well preserved.

In addition to the color accuracy, the previous arts are all limited to the color transformation.
The \textit{eyebrows shape} is ignored in makeup transfer, which is actually crucial for many makeup types.
In contrast, the novel FAT model can naturally transfer the spatial attributes of the reference image.
It estimates and aligns the spatial attributes, and seamlessly applies distilled transformation on the source image. As shown in Fig~\ref{fig:intro}, it is also robust to position, color space, and expression inconsistency.

Following \cite{psgan}, we also provide a functionality comparison with representative methods in Tab.~\ref{tab:functionality}.
The proposed FAT is the \textbf{first makeup method that integrates color, texture, and shape transformation} in a single, unified model.

\subsection{Quantitative Comparisons}\label{sec:quantitative-comparison}

In addition to the qualitative comparisons demonstrating the better results produced by FAT, we further conduct quantitative experiments for exhaustive comparison.

\begin{figure}[tbp]
    \centering
    \includegraphics[width=0.75\linewidth]{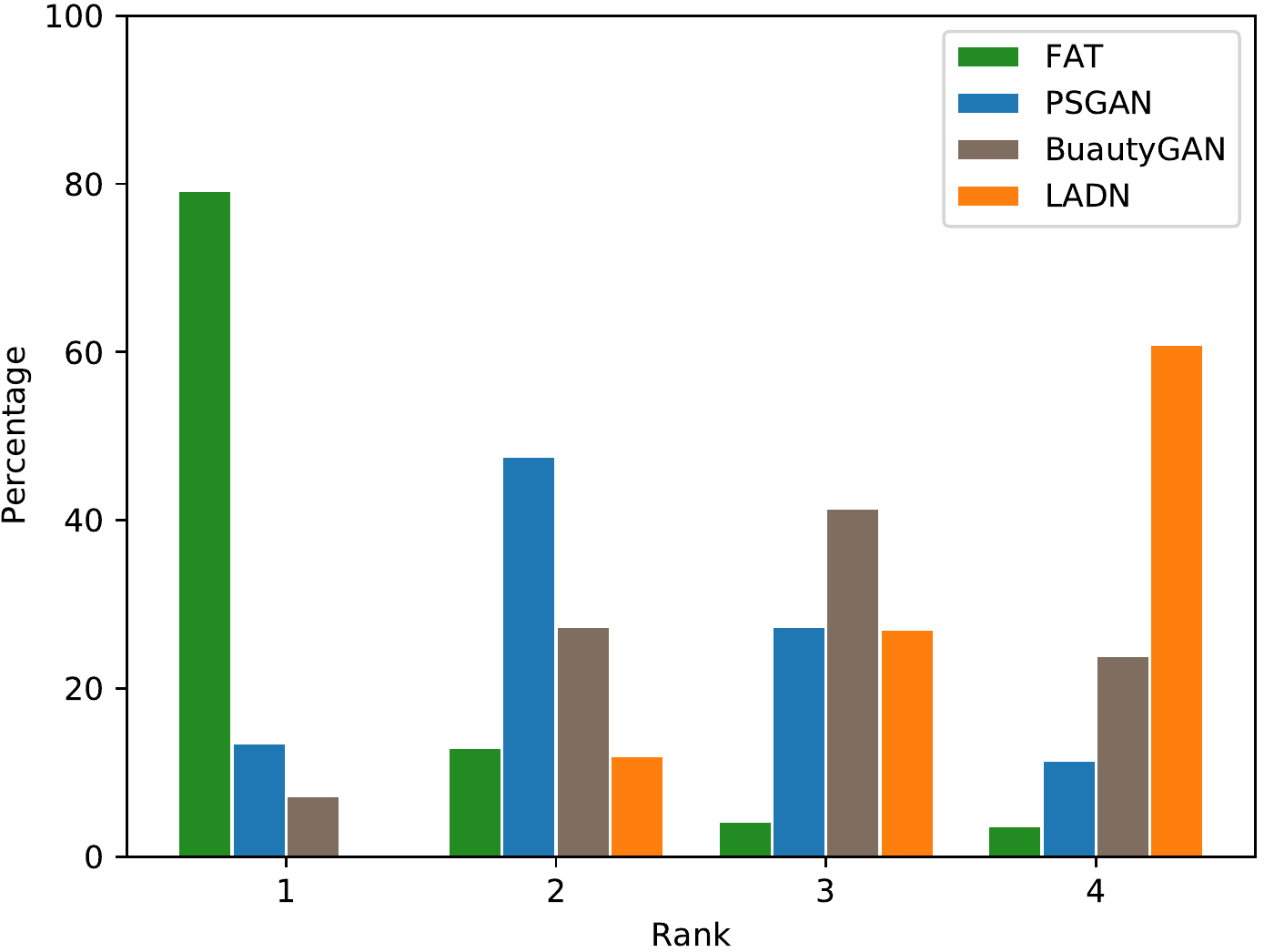}
    \caption{Blinded evaluation of methods by ranking.}
    \label{fig:blinded}
    \vspace{-0.2in}
\end{figure}
\noindent\textbf{Blinded Evaluation} First, we conduct a blinded evaluation to compare FAT with three state-of-the-art methods, BeautyGAN, LADN, and PSGAN, whose implementation and models are publicly available.
As PSGAN and FAT are trained with the MT-Dataset, we instead use the M-Wild dataset as the test set for a fair comparison.
We randomly sample source and reference images from M-Wild without replacement, resulting in 200 source-reference pairs. 
Each of these methods is fed with the image pairs to collect the results, and 10 volunteers are invited as testers to evaluate the generated images.

For each image pair, the generated results are randomly shuffled and testers are asked rank the results independently according to two standards: (1) How the generated faces are realistic that you can hardly distinguish it from real makeup faces; (2) How well is the makeup reconstructed that you may think it is the same makeup with the reference.
Then the images are ranked from the best to the worst subjectively by each tester.
In this experiment, we compare FAT instead of Spatial FAT, because Spatial FAT is the only method that can perform spatial transformation and will be easily recognized among others.

The rank distribution is shown in Fig.~\ref{fig:blinded}.
Due to the significant differences between the test data and the training data of LADN, its test performance is limited.
Verifying a better reconstruction of colors, the generated images from FAT are mostly chosen as the first rank.

\begin{figure}[tbp]
    \centering
    \includegraphics[width=0.95\linewidth]{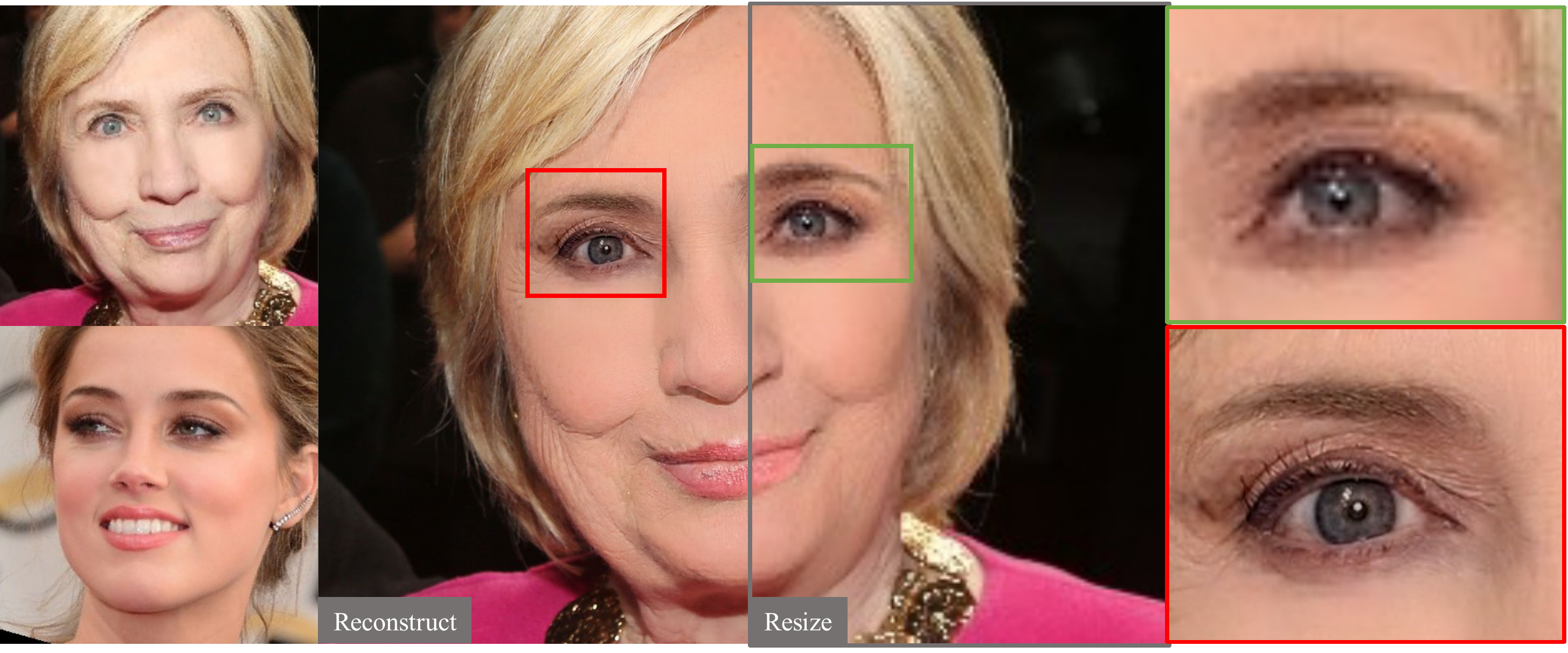}
    \caption{Post processing using pyramid reconstruction. Left: source and reference images; Middle: post-processing using resize and our reconstruction; Right: zoom-in view.}
    \label{fig:post-processing}
    \vspace{-0.1in}
\end{figure}

\noindent\textbf{Efficiency Evaluation}
\begin{table}[tb]
\centering
\begin{tabularx}{1.0\linewidth}{@{}l*{5}X@{}}
\toprule
\multirow{2}{*}{Method}     & \multicolumn{2}{c}{Inference Time}   &  \multirow{2}{*}{Attention}  \\
                            & CPU (s)  & GPU (ms) &  \\ \midrule
BeautyGAN                   & 1.15     &  9.3    &  $\times$ \\
LADN                        & 4.64     & 75.9    &  $\times$ \\
PSGAN                       & 3.02     & 146.6   &  $\checkmark$ \\
FAT                         & 1.93     & 22.0    &  $\checkmark$ \\
Spatial FAT                 & 2.16     & 26.4    &  $\checkmark$ \\
   \bottomrule
\end{tabularx}
\caption{Comparison of inference time (FPS). For fair comparison, we re-implement PSGAN and the reported speed is {\it actually} much faster than the original version.}
\label{tab:efficiency}
\vspace{-0.2in}
\end{table}
Besides the effectiveness, running efficiency is another major concern in real-world applications. 
Using the released implementations of the state-of-the-art methods, we compare their running efficiency with FAT.
The results are shown in Tab.~\ref{tab:efficiency}, where both FAT and Spatial FAT demonstrate highly competitive inference efficiency.
Benefiting from the paralleled design of attention, FAT is more than 5 times faster than its baseline PSGAN which has a static and sequential attention mechanism.

\subsection{High-Resolution Generation}\label{sec:post-processing}

Makeup transfer is usually performed on low-resolution images, e.g., 256$\times$256 for PSGAN~\cite{psgan} and 361$\times$361 for LADN~\cite{ladn}.
In consideration of the squared increasing of computational cost, we also use 256$\times$256 images as the input of the generator.
However, the lost of high-frequency signals can be alleviated via pyramid reconstruction~\cite{wiki:reconstruct}.
Given a high-resolution source image $\tilde{x}$, it is cropped around the face and resized to low-resolution input $x$.
The sampling loss from $\tilde{x}$ to $x$ can be partially recovered by measuring the deviation of interpolation as:
\begin{equation}
\begin{split}
    \tilde{z} &= \Delta + z\\
    \Delta &= \tilde{x} - interpolate(x),
\end{split}
\end{equation}
where $-$ and $+$ are element-wise operations, and $interpolate(\cdot)$ indicates bi-linear interpolation. 
The effect of pyramid reconstruction is shown in Fig.~\ref{fig:post-processing}, where the resolution of source image is $1200\times900$.
In our observation, the pyramid reconstruction is at least sufficient for 1k resolution facial images to produce natural results.

\section{Ablation and Generalization}\label{sec:ablation}

In addition to validation experiments, further experiments aiming at ablation and generalization study are provided in this section.

\subsection{Ablation Study}

Since FAT barely introduces hyper-parameters, we separately validate the effectiveness of each alternative to form our ablation experiments.

\noindent\textbf{Static Attention versus FAT}
\begin{figure}
    \vspace{-0.2in}
    \centering
    \includegraphics[width=\linewidth]{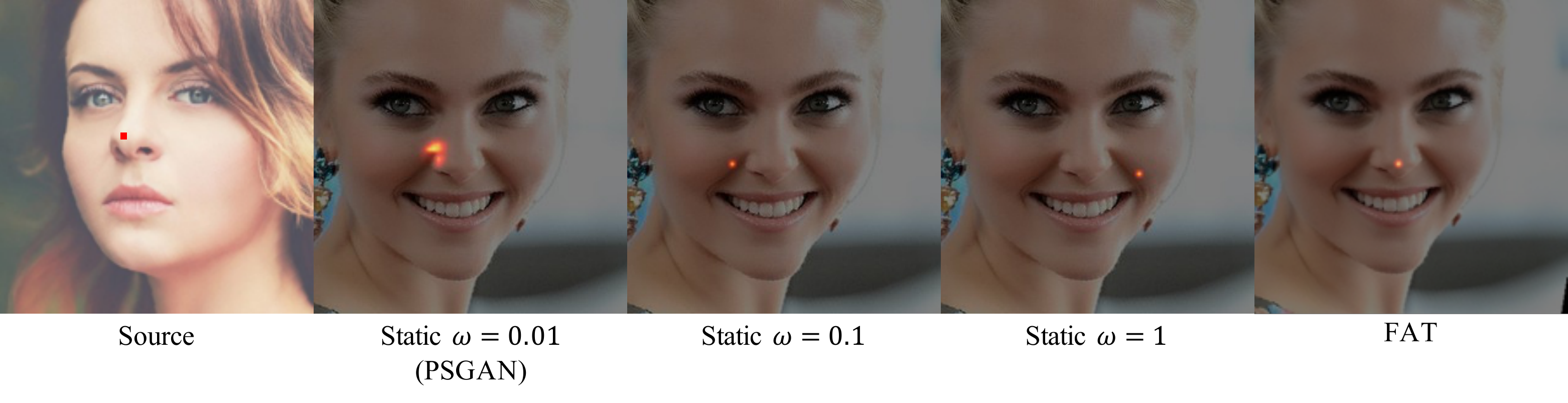}
    \caption{Visualization of the attention weights of static attention and FAT.
    The associated point at the source is marked with a red point.
    With visual features, the attention weight will be more focused. However, static dot-product attention may drift and fails to collaborate with visual features. Details are provided in Sec.~\ref{sec:ablation}.}
    \label{fig:ablation}
    \vspace{-0.1in}
\end{figure}
The dot-product attention inside FAT (Eq.~\ref{eq:atten}) can be used as an attention mechanism without optimizable parameters, which is the practice of \cite{psgan}.
As shown in the paper of \cite{psgan}, the static attention mainly relies on landmark embedding and must dilute the visual features with a small factor $\omega$ (set as 0.01 in their experiments):
\begin{equation}
\begin{split}
    A_{static}(\hat{x}, \hat{y}) &= Softmax(Sim(\hat{x},\hat{y})),\\
    Sim(\hat{x}, \hat{y}) &=  Concat(\omega\hat{x}, LE_x)^TConcat(\omega\hat{y}, LE_y),
\end{split}
\end{equation}
where $LE_x$ and $LE_y$ are landmark embeddings.
In contrast, our FAT mainly fully utilize visual features to precisely estimate the colors, and the landmark embedding is auxiliary for modeling spatial correspondence.
The attention weights of static dot-product attention and FAT is shown in Fig.~\ref{fig:ablation}.
Utilization of visual features leads to more concentrated attention, while drifting static attention positions.
With learnable weights, FAT adaptively locates the corresponding positions in the reference images, allowing precise transfer of facial attributes.

\noindent\textbf{Spatial Transformation without Spatial FAT}
\begin{figure}
    \centering
    \includegraphics[width=\linewidth]{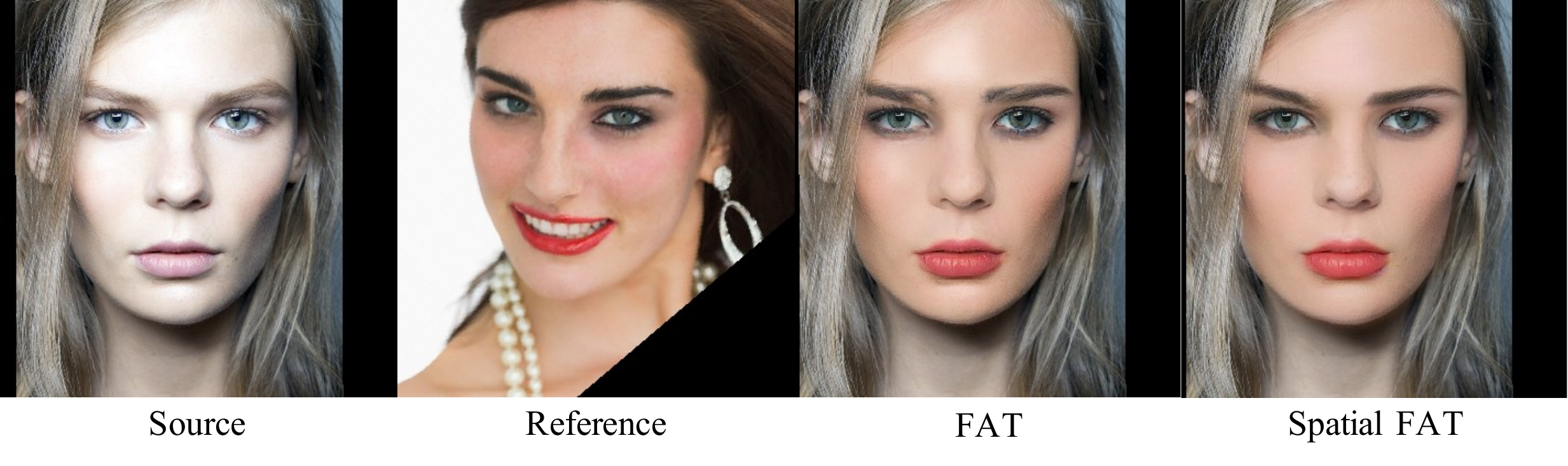}
    \caption{FAT and Spatial FAT under spatial supervision. Limited by the linear transformation function, FAT can hardly handle spatial transformation, where Spatial FAT performs better (notice the difference in the shape of the eyebrows).}
    \label{fig:ablation-spatial-fat}
    \vspace{-0.2in}
\end{figure}
As shown in Sec.~\ref{sec:label-generation}, the TPS-based ground truth generation is specially designed to provide guidance for Spatial FAT.
To verify the necessity of Spatial FAT over FAT, we directly adopt FAT for spatial transformation training.
Instead of its corresponding ground truth, we train FAT with the supervision for Spatial FAT where spatial transformation is applied in Fig.~\ref{fig:ablation-spatial-fat}.
Due to the linearly formulated transformation of FAT, it is theoretically not capable with spatial transformation. Thus, Spatial FAT is required for applications where spatial transformations are desired. Moreover, Spatial FAT achieves better color reconstruction benefiting from the extra spatial alignment that strengthens the correspondence.

\subsection{Generalization to Facial Age Transfer}\label{sec:age-transfer}
Enabling both color and spatial transformation, the proposed method can be generalized to a wide range of scenarios.
We conduct experiments on another popular face editing task, facial age editing to demonstrate the generalization of Spatial FAT.
The goal of facial age editing is to generate facial images at a different age of the source face.
In our experiments, we alternatively use a reference image to exhibit the targeted age, and the task is referred to as facial age transfer.
% transfer a young source face to the reference face with an aged style.
% In addition, our FAT framework performs well in other face editing tasks, such as facial age editing. Note that histogram matching ignores the spatial correspondences of the source and reference image, some facial features, such as wrinkle, could not be well adapted by histogram matching. In FAT framework, TPS makes face region alignment for the source and reference image. Benefit from the geometry alignment procedure, our FAT framework can also perform well on other facial editing tasks as a unified framework. 

We collect two groups of high-resolution images for training, images from the CelebAMask-HQ~\cite{DBLP:conf/iclr/KarrasALL18} dataset as junior faces, and senior faces generated by HRFAE\cite{hrfae}.
With the namely junior and senior group of images, age transfer using Spatial FAT is trained.
The same experiment is also conducted on our baseline PSGAN~\cite{psgan} with the same data for comparison.

\begin{figure}[tbp]
    \vspace{-0.2in}
    \centering
    \includegraphics[width=1.0\linewidth]{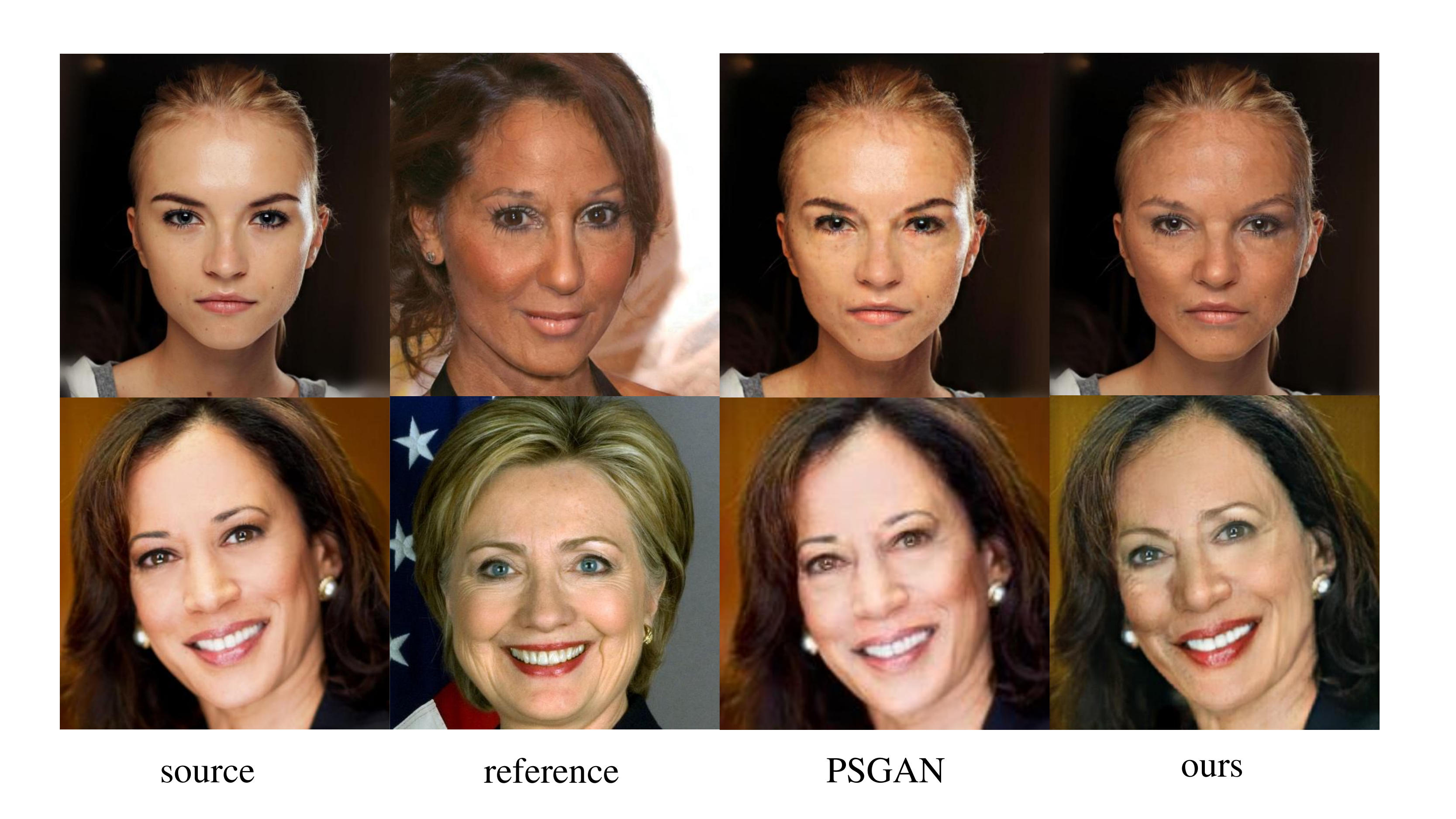}
    \caption{Facial age transfer using PSGAN and our Spatial FAT.}
    \label{fig:age}
    \vspace{-0.2in}
\end{figure}

As shown in Fig~\ref{fig:age}, Spatial FAT is also more effective in facial age transfer.
Limited by the formulation and transfer strategy, PSGAN transfers colors around eyes and eyebrows, but neglects detailed attributes that identify facial ages.
These attributes, e.g., wrinkles and facial quality, are well transferred by Spatial FAT.
Clearly Spatial FAT is able to lay a general framework for facial attributes transfer, with the support to both color and spatial transformation.

\section{Conclusion}
In this paper, we present Facial Attribute Transformers (FAT) for high-quality makeup transfer.
With an adaptive attention mechanism, FAT can precisely distill and ideally apply facial attributes from the reference face to the source face.
We further extend FAT with TPS transformation, thus creating Spatial FAT, whhich can transfer shape attributes.
Extensive experiments verified the consistent effectiveness and clear advantages of FATs, where precise attribute reconstruction, high-resolution generation, and competitive efficiency are achieved.
Moreover, the versatility of FAT as a general facial attribute transfer method is validated by additional experiments on facial age transfer.

\appendix
\section{Cycle-GAN Training in Makeup Transfer}
As stated in the paper, Cycle-GANs enable training with unpaired data and are the common practice in makeup transfer. Following we provide detailed training objectives of Cycle-GANs, in terms of makeup transfer.

\begin{figure}
    \centering
    \includegraphics[width=0.95\columnwidth]{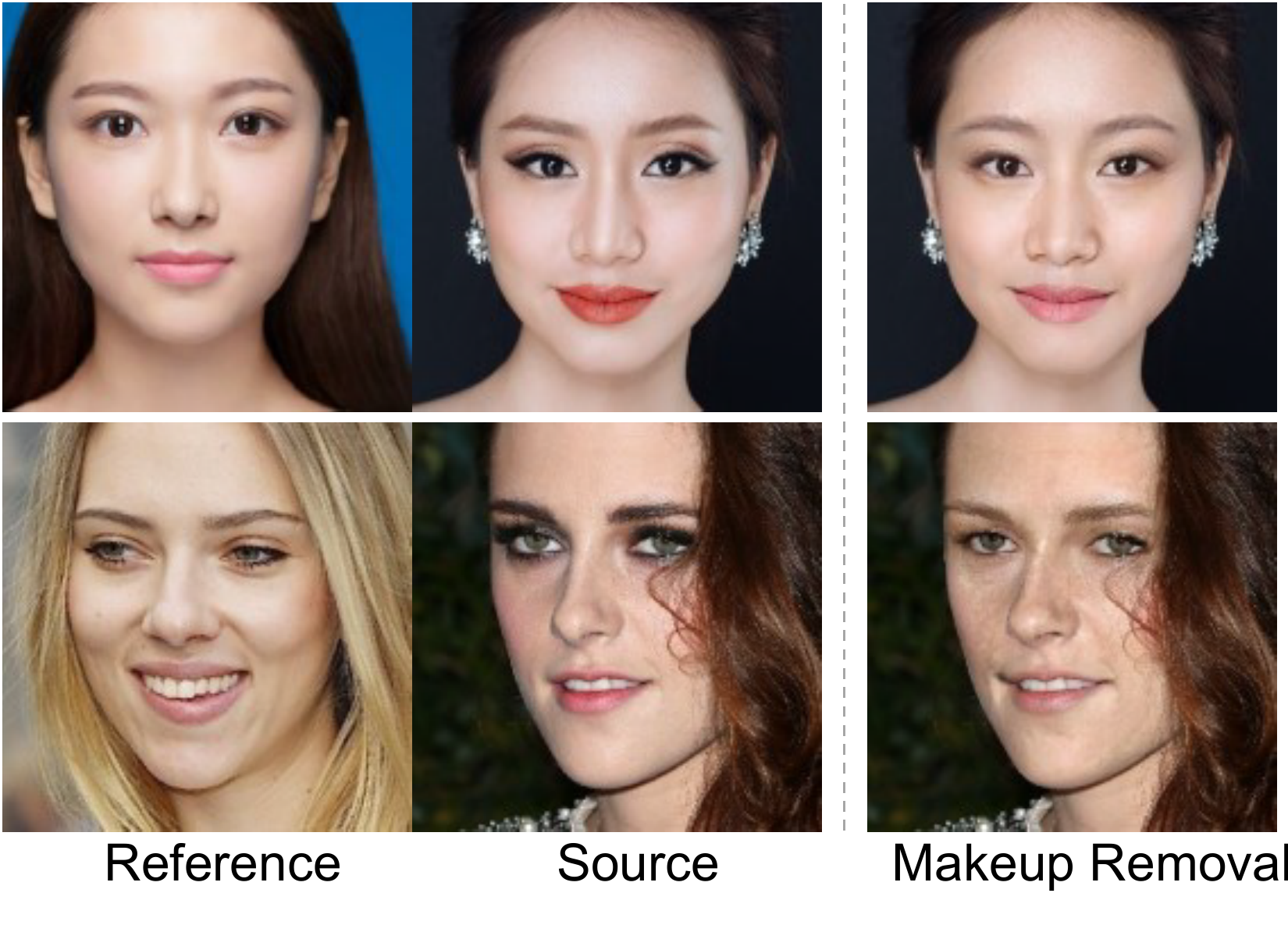}
    \caption{Makeup removal examples.}
    \label{fig:removal}
\end{figure}

We start from the aggregation of loss functions provided in the paper:
\begin{equation}\label{eq:loss}
    \begin{gathered}
    \begin{aligned} J_{D}^{adv} &= - \mathbb{E}_{x \sim \mathcal{P}_{X}}\left[\log D_{X}(x)\right] - \mathbb{E}_{y \sim \mathcal{P}_{Y}}\left[\log D_{Y}(y)\right] \\ &- \mathbb{E}_{x \sim \mathcal{P}_{X}, y \sim \mathcal{P}_{Y}}\left[\log \left(1-D_{X}(G(y, x))\right)\right] \\ &- \mathbb{E}_{x \sim \mathcal{P}_{X}, y \sim \mathcal{P}_{Y}}\left[\log \left(1-D_{Y}(G(x, y))\right)\right] 
        \end{aligned} \\
        \begin{aligned}
            J_G = \lambda_{adv} J_G^{adv} + \lambda_{cyc} J_G^{cyc} + \lambda_{per} J_G^{per} + \lambda_{make}J_G^{make},
        \end{aligned}
    \end{gathered} 
\end{equation}
where $J_D^{adv}$ and $J_G$ are the loss for discriminators and the generator, respectively.
Note that $J_D^{adv}$ is formed with two groups of terms, which corresponds two discriminators in makeup transfer:
One is for discriminating the generated makeup face from the reference, and the other is oriented at distinguishing the generated non-makeup face from the source.
The formulation indicates a clear message, that the removal of makeup is intrinsically trained along with the task of makeup transfer as shown in Fig.~\ref{fig:removal}.

Then let us look into the loss of the generator in Eq.~\ref{eq:loss}, where we add $J_G^{make}$ indicating makeup loss.
The adversarial loss of generator is given in the regular form of adversarial training:
\begin{equation}
    \begin{aligned} J_{G}^{adv} &= - \mathbb{E}_{x \sim \mathcal{P}_{X}, y \sim \mathcal{P}_{Y}}\left[\log D_{X}(G(y, x))\right] \\
    &- \mathbb{E}_{x \sim \mathcal{P}_{X}, y \sim \mathcal{P}_{Y}}\left[\log D_{Y}(G(x, y))\right]
    \end{aligned}
\end{equation}

The key that Cycle-GAN can be trained without paired data is laid in the cycle consistency loss~\cite{cyclegan}. L1 loss is used to supervise the reconstructed source from the generated face:
\begin{equation}
    \begin{aligned}
        J_G^{cyc} &=\mathbb{E}_{x \sim \mathcal{P}_{X}, y \sim \mathcal{P}_{Y}}\left[ \left\| G(G(x, y), x) - x \right\|_{1} \right]  \\ &+\mathbb{E}_{x \sim \mathcal{P}_{X}, y \sim \mathcal{P}_{Y}}\left[ \left\| G(G(y, x), y) - y \right\|_{1} \right].
    \end{aligned}
\end{equation}
For preserving the identity and perceptual details of source, A VGG-16 CNN is adopted to keep the consistency of extracted features. The perceptual loss is formulated as:
\begin{equation}
    \begin{aligned}
        J_G^{per} &=\mathbb{E}_{x \sim \mathcal{P}_{X}, y \sim \mathcal{P}_{Y}}\left[ \left\| F_l(G(x, y)) - F_l(x) \right\|_{2} \right]  \\ &+\mathbb{E}_{x \sim \mathcal{P}_{X}, y \sim \mathcal{P}_{Y}}\left[ \left\| F_l(G(y, x)) - F_l(y) \right\|_{2} \right],
    \end{aligned}
\end{equation}
where $F_l$ indicates the $l$th layer feature of the pre-trained VGG.

\section{Grid Generation of TPS}

Similar to STN~\cite{stn}, the Spatial FAT uses a grid generator to compute a sampling grid $\mathcal{P} = \{p_i\}$ on an image to form a transformation.
The 2D TPS transformation we introduce into Spatial FAT is parameterized by a $2\times(K+3)$ matrix~\cite{aster}:
\begin{equation}
    T = \begin{bmatrix}
    a_0, a_1, a_2, u \\
    b_0, b_1, b_2, v
    \end{bmatrix},
\end{equation}
where $u, v \in R^{1\times K}$.
\textbf{Following we use the formulation from ~\cite{aster} to describe the grid computation of 2D TPS.}
For a point $p \in R^{1 \times 2}$, its sampling point is computed by a linear projection:
\begin{equation}
    p' = T\begin{bmatrix}
    1\\
    p\\
    \phi(\|p-c_1\|)\\
    \vdots\\
    \phi(\|p-c_K\|),
    \end{bmatrix}
\end{equation}
where $\phi(r) = r^2log(r)$ as the radial basis kernel applied to the Euclidean distance between $p$ and control points $C$. Then the problem is to solve a linear system to find the coefficients of TPS:
\begin{equation}
    c_i' = T\begin{bmatrix}
    1\\
    p\\
    \phi(\|c_i-c_1\|)\\
    \vdots\\
    \phi(\|c_i-c_K\|),
    \end{bmatrix}, i,\dots,K,
\end{equation}
subject to boundary conditions:
\begin{equation}
\begin{aligned}
    0 &= u\textbf{1}\\
    0 &= v\textbf{1}\\
    0 &= uC_x^{T}\\
    0 &= vC_y^{T},
\end{aligned}
\end{equation}
where $C_x$ and $C_y$ are the first and second dimension coordinates, respectively. In a matrix form, $T$ has a closed-form solution.
\begin{equation}
    T = \begin{bmatrix}
    C', \textbf{0}^{2\times3}
    \end{bmatrix}\Delta_C^{-1}.
\end{equation}
Both the solving and application of TPS can be integrated into the neural networks, since they are differentiable matrix operations.

\section{Future Work}
As shown in the examples, the Spatial FAT demonstrates promising capability of transferring spatial attributes. Thus, further explorations on a wider spectrum of facial attributes transfer, such as facial expression transfer, are desired.

{\small
\bibliographystyle{ieee_fullname}
\bibliography{egbib}
}

\end{document}